%% file: emnlp2018.tex
\documentclass[11pt,a4paper]{article}
\usepackage[hyperref]{emnlp2018}
\usepackage{times}
\usepackage{latexsym}
\usepackage{url}
\usepackage{amsmath,amsthm,amssymb,amsfonts}
\usepackage{graphicx}
\usepackage{bbm}
\usepackage{algorithm, algorithmic}
\usepackage{multirow, rotating}
\usepackage{bm}
\usepackage{paralist}
\usepackage{subfigure}
\usepackage{enumitem}
\usepackage{todonotes}

\aclfinalcopy 

\newcommand{\nop}[1]{}
\newcommand{\textbox}[2]{{\color{#1}\fbox{\normalcolor#2}}}

\title{Subgoal Discovery for Hierarchical Dialogue Policy Learning}

\author{Da Tang$^{\star}$\quad Xiujun Li$^{\dagger}$\quad Jianfeng Gao$^{\dagger}$\quad Chong Wang$^{\ddag}$\quad Lihong Li$^{\ddag}$\quad Tony Jebara$^{\star}$ \\
  $^{\dagger}$Microsoft Research, Redmond, WA, USA \\
  $^{\star}$Columbia University, NY, USA \\
  $^{\ddag}$Google Inc., Kirkland, WA, USA \\
  {\tt \{xiul,jfgao\}@microsoft.com} \\
  {\tt \{datang,jebara\}@cs.columbia.edu} \\
  {\tt \{chongw,lihong\}@google.com}
}

\date{}

\begin{document}
\maketitle
\begin{abstract}
Developing agents to engage in complex goal-oriented dialogues is challenging partly because the main learning signals are very sparse in long conversations. 
In this paper, we propose a divide-and-conquer approach that discovers and exploits the hidden structure of the task to enable efficient policy learning. First, given successful example dialogues, we propose the \emph{Subgoal Discovery Network} (SDN) to divide a complex goal-oriented task into a set of simpler subgoals in an \emph{unsupervised} fashion. We then use these subgoals to learn a multi-level policy by hierarchical reinforcement learning.
We demonstrate our method by building a dialogue agent for the composite task of travel planning. Experiments with simulated and real users show that our approach performs competitively against a state-of-the-art method that requires human-defined subgoals. 
Moreover, we show that the learned subgoals are often human comprehensible.
\end{abstract}



\section{Introduction}
\label{sec:introduction}

Consider we want to plan a trip to a distant city using a dialogue agent. The agent must make choices at each leg, e.g., whether to fly or to drive, whether to book a hotel. Each of these steps in turn involves making a sequence of decisions all the way down to \emph{lower}-level actions. For example, to book a hotel involves identifying the location, specifying the check-in date and time, and negotiating the price etc.  

The above process of the agent has a natural hierarchy: a top-level process selects which subgoal to complete, and a low-level process chooses primitive actions to accomplish the selected subgoal. Within the reinforcement learning (RL) paradigm, such a hierarchical decision making process can be formulated in the \emph{options} framework \cite{sutton1999between}, where subgoals with their own reward functions are used to learn policies for achieving these subgoals. These learned policies are then used as temporally extended actions, or options, for solving the entire task.

Based on the options framework, researchers have developed dialogue agents for complex tasks, such as travel planning, using hierarchical reinforcement learning (HRL)~\citep{cuayahuitl10evaluation}.  Recently, \citet{peng2017composite} showed that the use of subgoals mitigates the reward sparsity and leads to more effective exploration for dialogue policy learning. However, these subgoals need to be human-defined which limits the applicability of the approach in practice because the domain knowledge required to properly define subgoals is often not available in many cases.

In this paper, we propose a simple yet effective \emph{Subgoal Discovery Network} (SDN) that discovers useful subgoals automatically for an RL-based dialogue agent. 
The SDN takes as input a collection of successful conversations, and identifies ``hub'' states as subgoals. Intuitively, a hub state is a region in the agent's state space that the agent tends to visit frequently on successful paths to a goal but not on unsuccessful paths. Given the discovered subgoals, HRL can be applied to learn a hierarchical dialogue policy which consists of (1) a top-level policy that selects among subgoals, and (2) a low-level policy that chooses primitive actions to achieve selected subgoals. 


We present the first study of learning dialogue agents with automatically discovered subgoals. We demonstrate the effectiveness of our approach by building a composite task-completion dialogue agent for travel planning. Experiments with both simulated and real users show that an agent learned with discovered subgoals performs competitively against an agent learned using expert-defined subgoals, and significantly outperforms an agent learned without subgoals. We also find that the subgoals discovered by SDN are often human comprehensible.


\section{Background}
\label{sec:backgroun}
A goal-oriented dialogue can be formulated as a Markov decision process, or MDP~\cite{levin00stochastic}, 
in which the agent interacts with its environment over a sequence of discrete steps. At each step $t\in\{0,1,\ldots\}$, the agent observes the current state $s_t$ of the conversation~\cite{henderson15machine,mrksic17neural,li2017end}, and chooses action $a_t$ according to a policy $\pi$. Here, the action may be a natural-language sentence or a speech act, among others. Then, the agent receives a numerical reward $r_t$ and switches to next state $s_{t+1}$. The process repeats until the dialogue terminates. The agent is to learn to choose \emph{optimal} actions $\{a_t\}_{t=1,2,\ldots}$ so as to maximize the total \emph{discounted} reward 
$r_0+\gamma r_1+\gamma^2 r_2 + \cdots $,
where $\gamma \in [0,1]$ is a discount factor. This learning paradigm is known as reinforcement learning, or RL~\cite{Sutton98Reinforcement}.

When facing a complex task, it is often more efficient to divide it into multiple simpler sub-tasks, solve them, and combine the partial solutions into a full solution for the original task. Such an approach may be formalized as hierarchical RL (HRL) in the options framework~\cite{sutton1999between}. An option can be understood as a subgoal, which consists of an initiation condition (when the subgoal can be triggered), an option policy to solve the subgoal, and a termination condition (when the subgoal is considered finished).

When subgoals are given, there exist effective RL algorithms to learn a hierarchical policy. A major open challenge is the automatic discovery of subgoals from data, the main innovation of this work is covered in the next section.


\section{Subgoal Discovery for HRL}
\begin{figure}[ht!]
\centering
\includegraphics[clip=true, trim=0.0cm 7.9cm 7.5cm 0.0cm, width=1\linewidth, scale=1.0]{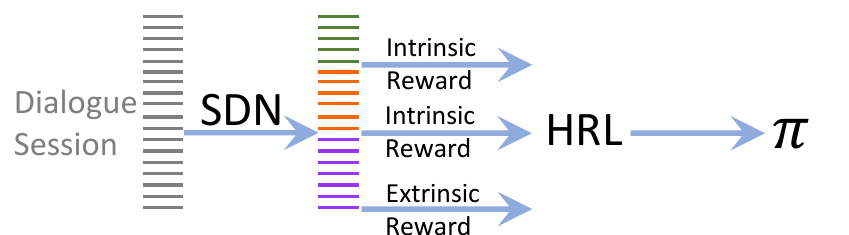}
\vspace{-3mm}
\caption{The workflow for HRL with subgoal discovery. In addition to the extrinsic reward at the end of the dialogue session, HRL also uses intrinsic rewards induced by the subgoals (or the ends of dialogue segments). Section~\ref{sec:policy_learning} details the reward design for HRL.}
\label{fig:dialogue_flows}
\end{figure}

Figure~\ref{fig:dialogue_flows} shows the overall workflow of our proposed method of using automatic subgoal discovery for HRL. First a dialogue session is divided into several segments. Then at the end of those segments (subgoals), we equip an intrinsic or extrinsic reward for the HRL algorithm to learn a hierarchical dialogue policy. Note that only the last segment has an extrinsic reward. The details of the segmentation algorithm and how to use subgoals for HRL are presented in Section~\ref{sec:subgoal_discovery} and Section~\ref{sec:HRL_SDN}.

\subsection{Subgoal Discovery Network}
\label{sec:subgoal_discovery}
\begin{figure}[ht!]
\centering
\includegraphics[clip=true, trim=0.2cm 7.0cm 1cm 5cm, width=1\linewidth, scale=1.0]{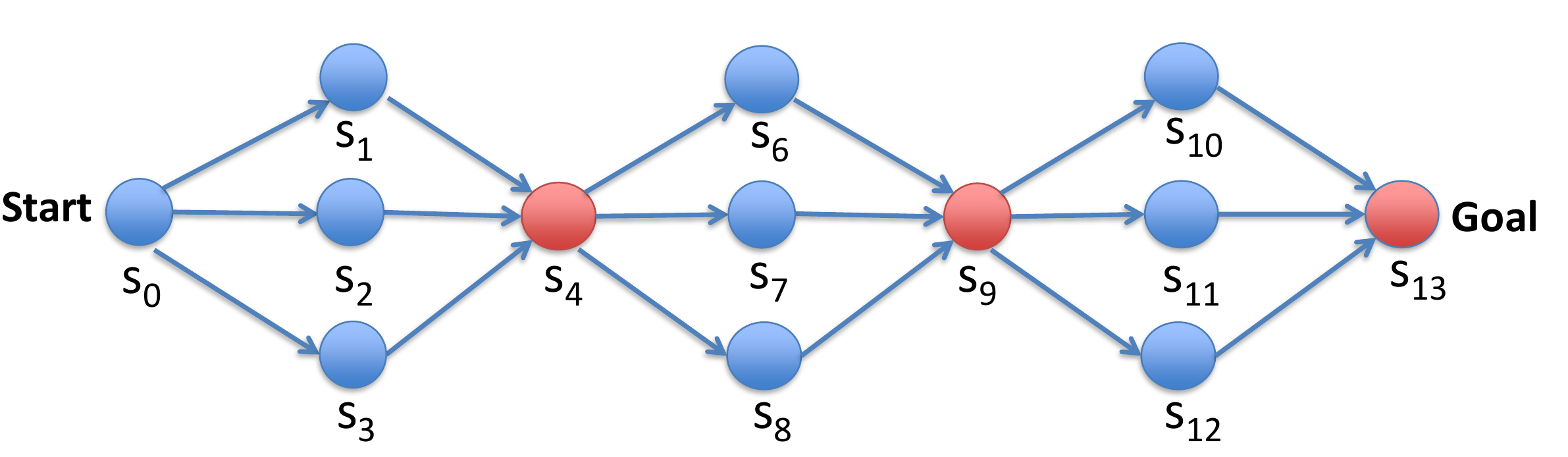}
\vspace{-5mm}
\caption{Illustration of ``subgoals''. Assuming that there are three state trajectories $(s_0, s_1, s_4, s_6, s_9, s_{10}, s_{13})$, $(s_0, s_2, s_4, s_7, s_9, s_{11}, s_{13})$ and $(s_0, s_3, s_4, s_8, s_9, s_{12}, s_{13})$. Then red states $s_4$, $s_9$, $s_{13}$ could be good candidates for ``subgoals''.
\label{fig:paths}}
\vspace{-1mm}
\end{figure}

Assume that we have collected a set of successful state trajectories of a task, as shown in Figure~\ref{fig:paths}. We want to find subgoal states, such as the three red states $s_4$, $s_9$ and $s_{13}$, which form the ``hubs'' of these trajectories. These hub states indicate the subgoals, and thus divide a state trajectory into several segments, each for an option\footnote{There are many ways of creating a new option $\langle I,\pi,\beta \rangle$ for a discovered subgoal state. For example, when a subgoal state is identified at time step $t$, we add to $I$ the set of states visited by the agent from time $t-n$ to $t$, where $n$ is a pre-set parameter. $I$ is therefore the union of all such states over all the state trajectories. The termination condition $\beta$ is set to 1 when the subgoal is reached or when the agent is no longer in $I$, and to $0$ otherwise. In the deep RL setting where states are represented by continuous vectors, $\beta$ is a probability whose value is proportional to the vector distance e.g., between current state and subgoal state.}. 

Thus, discovering subgoals by identifying hubs in state trajectories is equivalent to segmenting state trajectories into options. 
In this work, we formulate subgoal discovery as a state trajectory segmentation problem, and address it using the Subgoal Discovery Network (SDN), inspired by the sequence segmentation model~\cite{wang2017sequence}.

\begin{figure}[t]
\centering
\includegraphics[clip=true, trim=1cm 3.0cm 2cm 0cm, width=1\linewidth, scale=1.0]{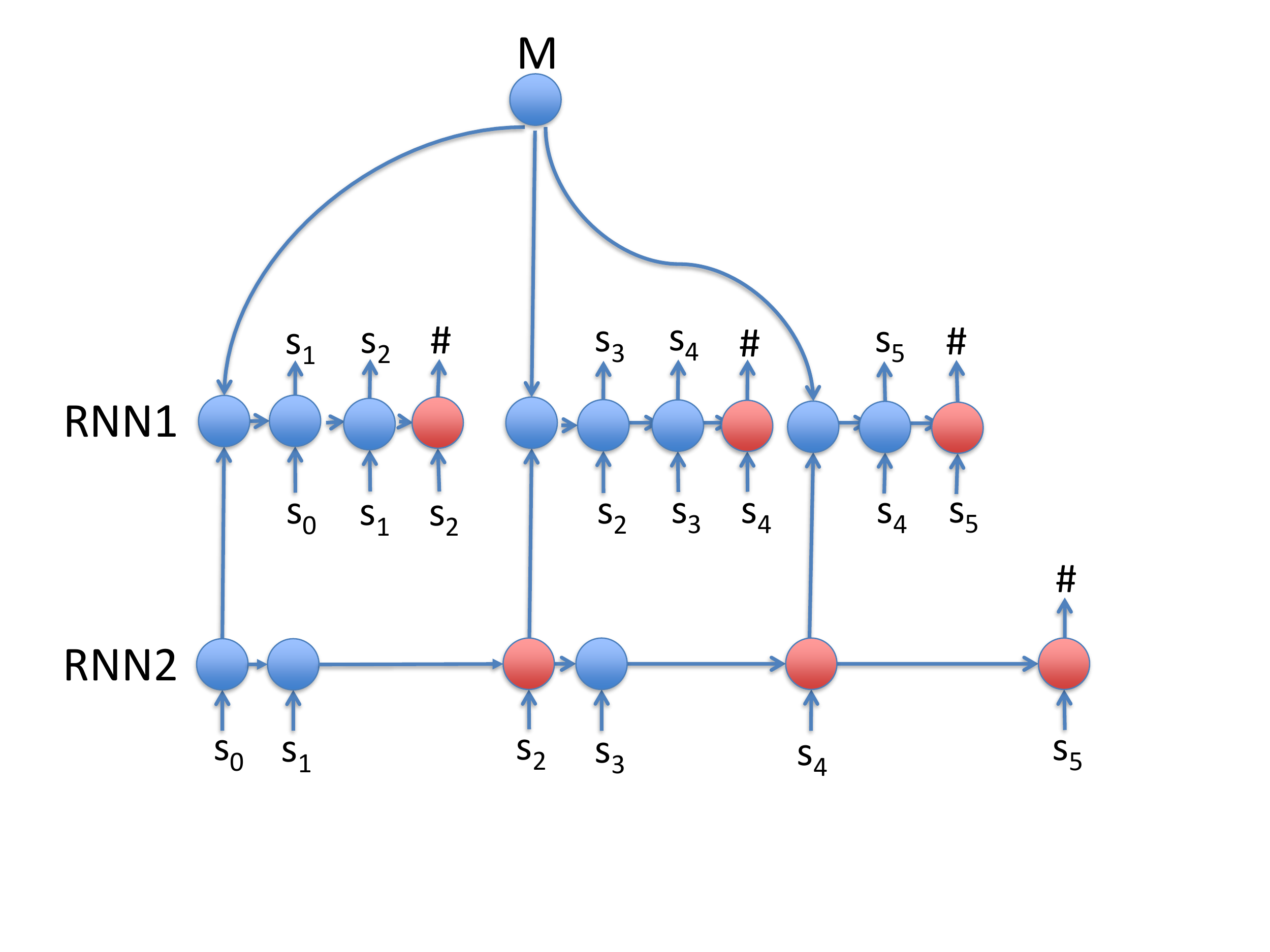}
\vspace{-5mm}
\caption{Illustration of SDN for state trajectory $(s_0,\ldots,s_5)$ with $s_2$, $s_4$ and $s_5$ as subgoals. Symbol \# is the termination. The top-level RNN (RNN1) models segments and the low-level RNN (RNN2) provides information about previous states from RNN1. The embedding matrix $M$ maps the outputs of RNN2 to low dimensional representations so as to be consistent with the input dimensionality of RNN1. Note that state $s_5$ is associated with two termination symbols \#; one is for the termination of the last segment and the other is for the termination of the entire trajectory.}
\label{fig:rnn}
\end{figure}

\paragraph{The SDN architecture.} SDN repeats a two-stage process of generating a state trajectory segment, until a trajectory termination symbol is generated: first it uses an initial segment hidden state to start a new segment, or a trajectory termination symbol to terminate the trajectory, given all previous states; if the trajectory is not terminated, then keep generating the next state in this trajectory segment given previous states\nop{in this segment} until a segment termination symbol is generated. We illustrated this process in Figure~\ref{fig:rnn}. 

We model the likelihood of each segment using an RNN, denoted as RNN1. During the training, at each time step, RNN1 predicts the next state with the current state as input, until it reaches the option termination symbol \#. Since different options are under different conditions, it is not plausible to apply a fixed initial input to each segment. 
Therefore, we use another RNN (RNN2) to encode all previous states to provide relevant information and we transform these information to low dimensional representations as the initial inputs for the RNN1 instances. This is based on the \emph{causality} assumption of the options framework~\cite{sutton1999between} --- the agent should be able to determine the next option given all previous information, and this should not depend on information related to any later state. The low dimensional representations are obtained via a global subgoal embedding matrix $M\in\mathbb R^{d\times D}$, where $d$ and $D$ are the dimensionality of RNN1's input layer and RNN2's output layer, respectively. Mathematically, if the output of RNN2 at time step $t$ is $o_t$, then from time $t$ the RNN1 instance has $M\cdot {\mathrm{softmax}}(o_t)$ as its initial input\footnote{${\mathrm{softmax}}( o_t)_i=\exp(o_{t,i})/\sum\limits_{i'=1}^D\exp(o_{t,i'})\in\mathbb R^D$ for $o_t=(o_{t,1},\ldots,o_{t,D})$.
}. $D$ is the number of subgoals we aim to learn. Ideally, the vector ${\mathrm{softmax}}(o_t)$ in a well-trained SDN is close to an one-hot vector. Therefore, $M\cdot {\mathrm{softmax}}(o_t)$ should be close to one column in $M$ and we can view that $M$ provides at most $D$ different ``embedding vectors'' for RNN1 as inputs, \nop{This design makes most of the RNN1 initial inputs, which control the generative process of the state transition sequences outputted by the corresponding RNN1 instances, fall into at most $D$ small regions,}indicating at most $D$ different subgoals. Even in the case where ${\mathrm{softmax}}(o_t)$ is not close to any one-hot vector, choosing a small $D$ helps avoid overfitting.

\paragraph{Segmentation likelihood.} Given the state trajectory $(s_0,\ldots,s_5)$, assuming that $s_2$, $s_4$ and $s_5$ are the discovered subgoal states, we model the conditional likelihood of a proposed segmentation $\sigma=((s_0,s_1,s_2),(s_2,s_3,s_4),(s_4,s_5))$ as $p(\sigma|s_0)=p((s_0,s_1,s_2)|s_0)\cdot p((s_2,s_3,s_4)|s_{0:2})\cdot p((s_4,s_5)|s_{0:4})$, where each probability term $p(\cdot|s_{0:i})$ is based on an RNN1 instance. And for the whole trajectory $(s_0,\ldots,s_5)$, its likelihood is the sum over all possible segmentations. 

Generally, for state trajectory $\bm{s}=(s_0,\ldots,s_T)$, we model its likelihood as follows\footnote{For notation convenience, we include $s_0$ into the observational sequence, though $s_0$ is always conditioned upon.}:
\begin{equation}
L_S(\bm s)=\sum\limits_{\sigma\subseteq\mathcal S(\bm s), \text{length}(\sigma)\le S}\prod\limits_{i=1}^{\text{length}(\sigma)}p(\sigma_i|\tau(\sigma_{1:i})),
\label{eqn:likeli}
\end{equation}
where $\mathcal S(\bm s)$ is the set of all possible segmentations for the trajectory $\bm s$, $\sigma_i$ denotes the $i^\text{th}$ segment in the segmentation $\sigma$, and $\tau$ is the concatenation operator. $S$ is an upper limit on the maximal number of segments. This parameter is important for learning subgoals in our setting since we usually prefer a small number of subgoals. This is different from~\citet{wang2017sequence}, where a maximum segment length is enforced.

We use maximum likelihood estimation with Eq.~\eqref{eqn:likeli} for training. However, the number of possible segmentations is exponential in $\mathcal S(\bm s)$ and the naive enumeration is intractable. Here, dynamic programming is employed to compute the likelihood in Eq.~\eqref{eqn:likeli} efficiently: for a trajectory $\bm s=(s_0,\ldots,s_T)$, if we denote the sub-trajectory $(s_i,\ldots,s_t)$ of $\bm s$ as $\bm s_{i:t}$, then its likelihood follows the below recursion:
\begin{equation*}
L_m(\bm s_{0:t})=
\begin{cases}
\sum\limits_{i=0}^{t-1}L_{m-1}(\bm s_{0:i})p(\bm s_{i:t}|\bm s_{0:i}),&m>0,\\
I[t=0],&m=0.
\end{cases}
\end{equation*}
Here, $L_m(\bm s_{0:t})$ denotes the likelihood of sub-trajectory $\bm s_{0:t}$ with no more than $m$ segments and $I[\cdot]$ is an indicator function. $p(\bm s_{i:t}|\bm s_{0:i})$ is the likelihood segment $\bm s_{i:t}$ given the previous history, where RNN1 models the segment and RNN2 models the history as shown in Figure~\ref{fig:rnn}. With this recursion, we can compute the likelihood $L_S(\bm s)$ for the trajectory $\bm s=(s_0,\ldots,s_T)$ in $O(ST^2)$ time.

\paragraph{Learning algorithm.} We denote $\theta^s$ as the model parameter including the parameters of the embedding matrix $M$, RNN1 and RNN2. We then parameterize the segment likelihood function as $p(\bm s_{i:t}|\bm s_{0:i})=p(\bm s_{i:t}|\bm s_{0:i};\theta^s)$, and the trajectory likelihood function as $L_m(\bm s_{0:t})=L_m(\bm s_{0:t};\theta^s)$. 

%
Given a set of $N$ state trajectories $(\bm s^{(1)}, \ldots, \bm s^{(N)})$, we optimize $\theta^s$ by minimizing the negative mean log-likelihood with $L_2$ regularization term $\frac12\lambda||\theta^s||^2$ where $\lambda>0$, using stochastic gradient descent:
\begin{equation}
\small
\begin{aligned}
\mathcal L_S(\theta^s,\lambda)=-\frac1N\sum\limits_{i=1}^N\log L_S(\bm s^{(i)},\theta^s)+\frac12\lambda||\theta^s||^2.
\end{aligned}
\label{eqn:obj}
\end{equation}
Algorithm~\ref{algo:subgoal_discovery} outlines the training procedure for SDN using stochastic gradient descent.

\begin{algorithm}[htbp]
\small
\caption{Learning SDN}
\begin{algorithmic}[1]
\REQUIRE A set of state trajectories $(\bm s_1, \ldots \bm s_N)$, the number of segments limit $S$, initial learning rate $\eta>0$.
\STATE Initialize the SDN parameter $\theta^s$.
\WHILE {not converged}
\STATE Compute the gradient $\nabla_{\theta^s}\mathcal L_S(\theta^s,\lambda)$ of the loss $\mathcal L_S(\theta^s,\lambda)$ as in Eq. \eqref{eqn:obj}. 
\STATE Update $\theta^s\leftarrow\theta^s-\eta\nabla_{\theta^s}\mathcal L_S(\theta^s,\lambda)$.
\STATE Update the learning rate $\eta$.
\ENDWHILE
\end{algorithmic}
\label{algo:subgoal_discovery}
\end{algorithm}

\input{section_hrl}

\subsection{Hierarchical Policy Learning with SDN}
\label{sec:HRL_SDN}
We use a trained SDN in HRL as follows. The agent starts from the initial state $s_0$, keeps sampling the output from the distribution related to the top-level RNN (RNN1) until a termination symbol \# is generated, which indicates the agent reaches a subgoal. In this process, intrinsic rewards are generated as specified in the previous subsection. After \# is generated, the agent selects a new option, and repeats this process.

This type of naive sampling may allow the option to terminate at some places with a low probability. To stabilize the HRL training, we introduce a threshold $p \in (0, 1)$, which directs the agent to terminate an option if and only if the probability of outputting \# is at least $p$. We found this modification leads to better behavior of the HRL agent than the naive sampling method, since it normally has a smaller variance.

In the HRL training, the agent only uses the probability of outputting \# to decide subgoal termination. Algorithm~\ref{algo:hrl_training_subgoals} outlines the full procedure of one episode for hierarchical dialogue policies with a trained SDN in the composite task-completion dialogue system.

\section{Experiments and Results}
\label{sec:experiment}

We evaluate the proposed model on a travel planning scenario for composite task-oriented dialogues~\cite{peng2017composite}. Over the exchange of a conversation, the agent gathers information about the user's intent before booking a trip. The environment then assesses a binary outcome (success or failure) at the end of the conversation, based on (1) whether a trip is booked, and (2) whether the trip satisfies the user's constraints.

\begin{algorithm}[ht!]
\small
\caption{HRL episode with a trained SDN}
\begin{algorithmic}[1]
\REQUIRE A trained SDN $\mathcal M$, initial state $s_0$ of an episode, threshold $p$, the HRL agent $\mathcal A$.
\STATE Initialize an RNN2 instance $R_2$ with parameters from $\mathcal M$ and $s_0$ as the initial input.
\STATE Initialize an RNN1 instance $R_1$ with parameters from $\mathcal M$ and $M\cdot\text{softmax}(o^{\text{RNN2}}_0)$ as the initial input, where $M$ is the embedding matrix (from $\mathcal M$) and $o^{\text{RNN2}}_0$ is the initial output of $R_2$.
\STATE Current state $s\leftarrow s_0$.
\STATE Select an option $o$ using the agent $\mathcal A$.
\WHILE {Not reached the final goal}
\STATE Select an action $a$ according to $s$ and $o$ using the agent $\mathcal A$. Get the reward $r$ and the next state $s'$ from the environment.
\STATE Place $s'$ to $R_2$, denote $o^{\text{RNN2}}_t$ as $R_2$'s latest output and take  $M\cdot\text{softmax}(o^{\text{RNN2}}_t)$ as the $R_1$'s new input. Let $p_{s'}$ be the probability of outputting the termination symbol \#.
\IF {$p_{s'}\ge p$}
\STATE Select a new option $o$ using the agent $\mathcal A$.
\STATE Re-initialize $R_1$ using the latest output from $R_2$ and the embedding matrix $M$.
\ENDIF
\ENDWHILE
\end{algorithmic}
\label{algo:hrl_training_subgoals}
\end{algorithm}


\paragraph{Dataset.}
The raw dataset in our experiments is from a publicly available multi-domain dialogue corpus~\cite{elframes}. Following~\citet{peng2017composite}, a few changes were made to introduce dependencies among subtasks. For example, the hotel check-in date should be the same with the departure flight arrival date. The data was mainly used to create simulated users, and to build the knowledge bases for the subtasks of booking flights and reserving hotels.

\paragraph{User Simulator.}
In order to learn good policies, RL algorithms typically need an environment to interact with. In the dialogue research community, it is common to use simulated users for this purpose~\cite{schatzmann2007agenda,li2017end,liu2017iterative}. In this work, we adapted a publicly available user simulator~\cite{li2016user} to the composite task-completion dialogue setting with the dataset described above. 
During training, the simulator provides the agent with an (extrinsic) reward signal at the end of the dialogue. A dialogue is considered to be successful only when a travel plan is booked successfully, and the information provided by the agent satisfies user's constraints. 

\paragraph{Baseline Agents.}
We benchmarked the proposed agent (referred to as the \textit{m-HRL Agent}) against three baseline agents: 
\begin{itemize}[noitemsep,leftmargin=*,topsep=0pt]
\item A \textit{Rule Agent} uses a sophisticated, hand-crafted dialogue policy, which requests and informs a hand-picked subset of necessary slots, and then confirms with the user about the reserved trip before booking the flight and hotel.
\item A \textit{flat RL Agent} is trained with a standard deep reinforcement learning method, DQN~\cite{mnih2015human}, which learns a flat dialogue policy using extrinsic rewards only.
\item A \textit{h-HRL Agent} is trained with hierarchical deep reinforcement learning (HDQN), which learns a hierarchical dialogue policy based on human-defined subgoals~\cite{peng2017composite}. 
\end{itemize}

\begin{figure}[t!]
\centering
\includegraphics[width=\linewidth]{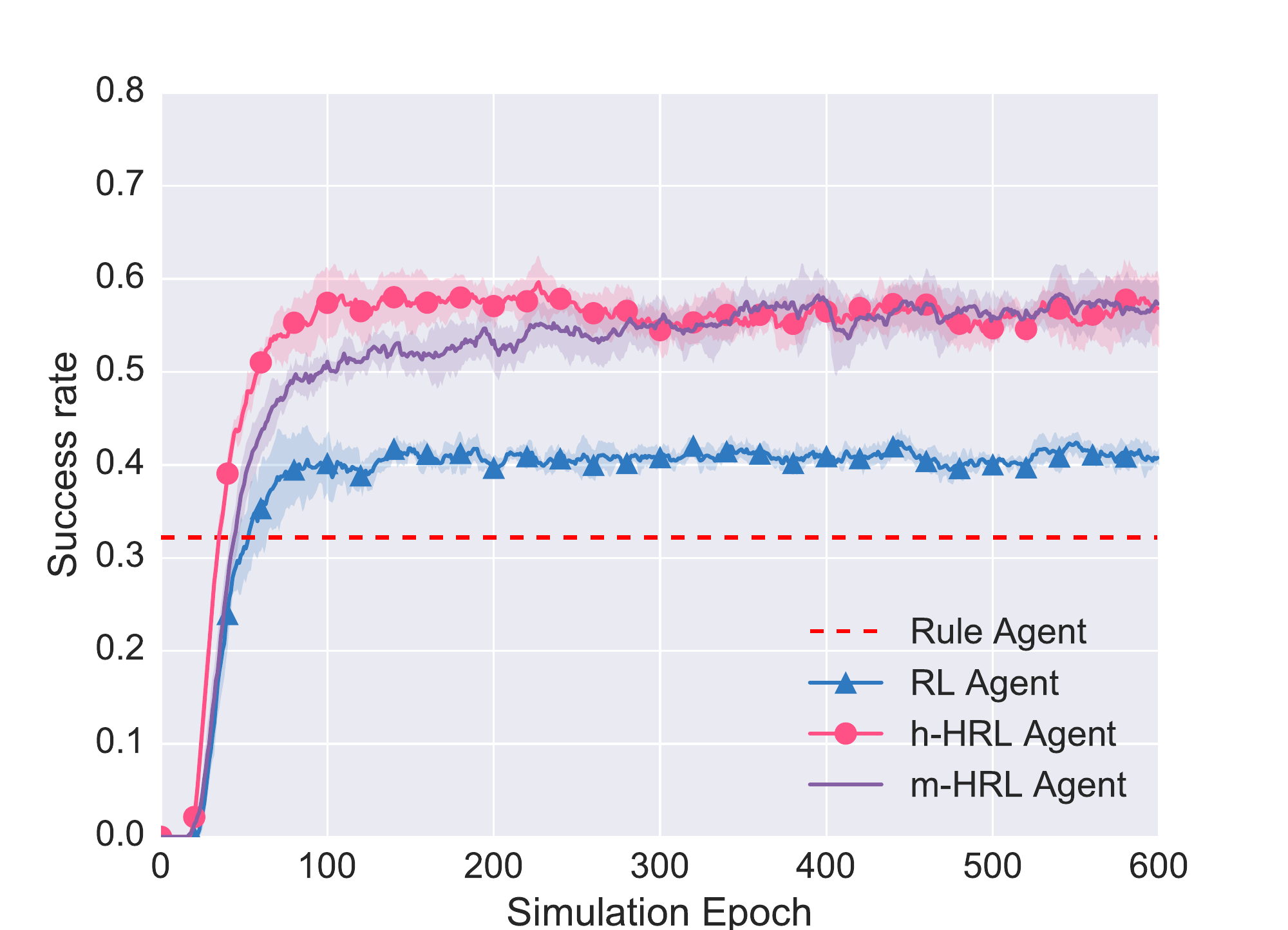}
\vspace{-5mm}
\caption{Learning curves of agents under simulation.}
\label{fig:sim_results}
\end{figure}

\paragraph{Collecting State Trajectories.}
Recall that our subgoal discovery approach takes as input a set of state trajectories which lead to successful outcomes. In practice, one can collect a large set of successful state trajectories, either by asking human experts to demonstrate (e.g., in a call center), or by rolling out a reasonably good policy (e.g., a policy designed by human experts).
In this paper, we obtain dialogue state trajectories from a rule-based agent which is handcrafted by a domain expert, the performance of this rule-based agent can achieve success rate of 32.2\% as shown in Figure~\ref{fig:sim_results} and Table~\ref{tab:results}. We only collect the successful dialogue sessions from the roll-outs of the rule-based agent, and try to learn the subgoals from these dialogue state trajectories.

\paragraph{Experiment Settings.}
To train SDN, we use RMSProp~\cite{tieleman2012lecture} to optimize the model parameters. For both RNN1 and RNN2, we use LSTM~\cite{hochreiter1997long} as hidden units and set the hidden size to $50$. We set embedding matrix $M$ with $D=4$ columns. As we discussed in Section~\ref{sec:subgoal_discovery}, $D$ captures the maximum number of subgoals that the model is expected to learn.
Again, to avoid SDN from learning many unnecessary subgoals, we only allow segmentation with at most $S=4$ segments during subgoal training.
The values for $D$ and $S$ are usually set to be a little bit larger than the expected number of subgoals (e.g., $2$ or $3$ for this task) since we expect a great proportion of the subgoals that SDN learns are useful, but not necessary for all of them. As long as SDN discovers useful subgoals that guide the agent to learn policies faster, it is beneficial for HRL training, even if some non-perfect subgoals are found. During the HRL training, we use the learned SDN to propose subgoal-completion queries. In our experiment, we set the maximum turn $L=60$.


We collected $N=1634$ successful, but imperfect, dialogue episodes from the rule-based agent in Table~\ref{tab:results} and randomly choose $80\%$ of these dialogue state trajectories for training SDN. The remaining $20\%$ were used as a validation set.

As illustrated in Section~\ref{sec:HRL_SDN}, SDN starts a new RNN1 instance and issues a subgoal-completion query when the probability of outputting the termination symbol \# is above a certain threshold $p$ (as in Algorithm~\ref{algo:hrl_training_subgoals}). In our experiment, $p$ is set to be 0.2, which was manually picked according to the termination probability during SDN training.

In dialogue policy learning, for the baseline RL agent, we set the size of the hidden layer to $80$. For the HRL agents, both top-level and low-level dialogue policies have a hidden layer size of $80$. RMSprop was applied to optimize the parameters. We set the batch size to be $16$. During training, we used $\epsilon$-greedy strategy for exploration with annealing and set $\gamma=0.95$. For each simulation epoch, we simulated $100$ dialogues and stored these state transition tuples in the experience replay buffers. At the end of each simulation epoch, the model was updated with all the transition tuples in the buffers in a batch manner.



\begin{table}[t!]
\begin{center}
\begin{tabular}{cccc}
\\ \hline
Agent & Success Rate & Turns & Reward \\ \hline 
Rule & .3220 & 46.23 & -24.02 \\ 
RL & .4440 & 45.50 & -1.834 \\ 
h-HRL & .6485 & 44.23 & 35.32 \\ 
m-HRL & .6455 & 44.85 & 34.77 \\ 
\hline
\end{tabular}
\end{center}
\caption{Performance of agents with simulated user.}
\label{tab:results}
\end{table}

\subsection{Simulated User Evaluation}
\label{sec:sim_user_eval}
In the composite task-completion dialogue scenario, we compared the proposed \textit{m-HRL} agent with three baseline agents 
in terms of three metrics: success rate\footnote{Success rate is the fraction of dialogues which accomplished the task successfully within the maximum turns.}, average rewards and average turns per dialogue session.

Figure~\ref{fig:sim_results} shows the learning curves of all four agents trained against the simulated user. Each learning curve was averaged over $5$ runs. Table~\ref{tab:results} shows the test performance where each number was averaged over $5$ runs and each run generated $2000$ simulated dialogues. We find that the HRL agents generated higher success rates and needed fewer conversation turns to achieve the users' goals than the rule-based agent and the flat RL agent. The performance of the m-HRL agent is tied with that of the h-HRL agent, even though the latter requires high-quality subgoals designed by human experts. 

\begin{figure}[t!]
\centering
\includegraphics[width=\linewidth]{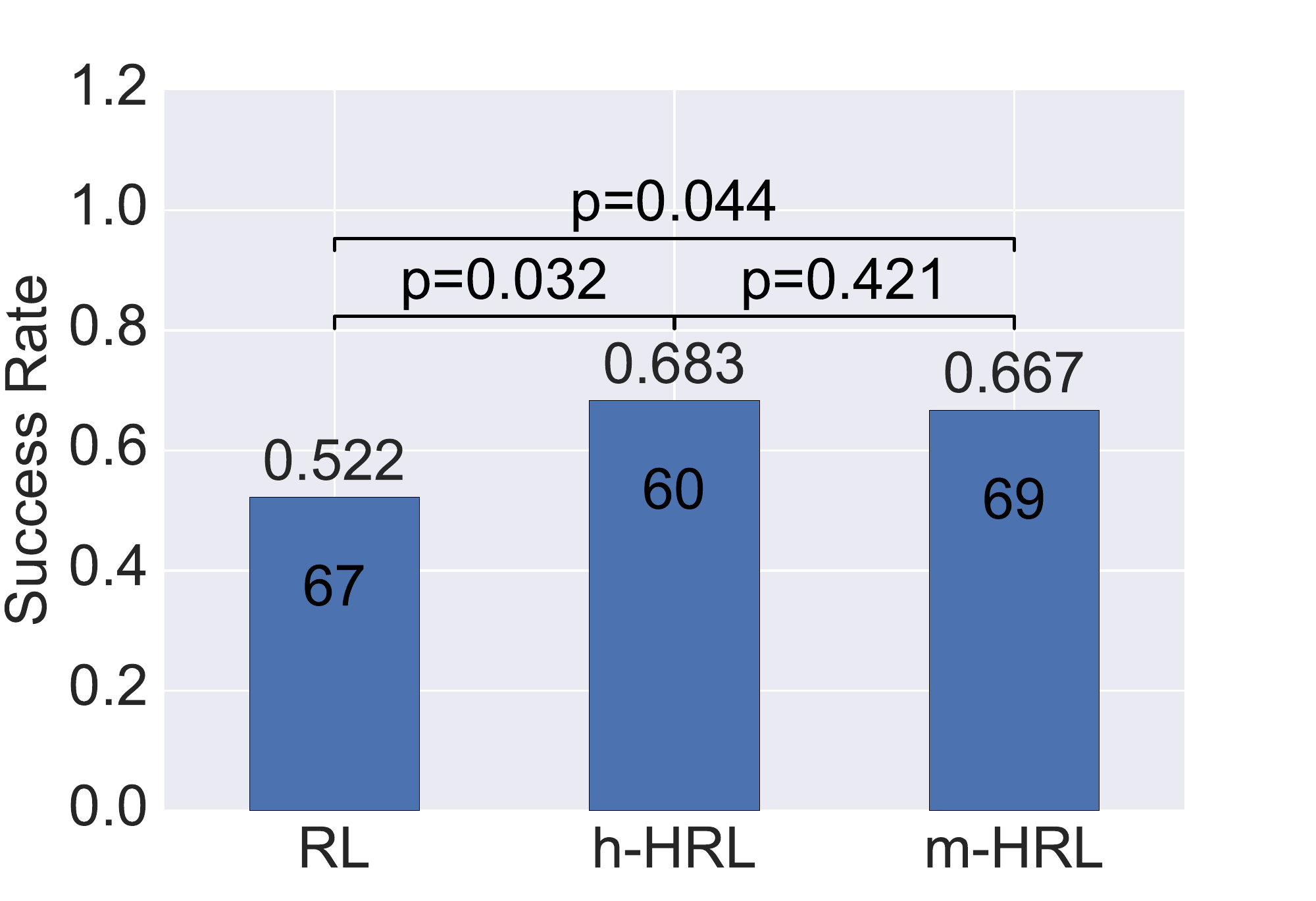}
\vspace{-5mm}
\caption{Performance of three agents tested with real users: success rate, number of dialogues and p-value are indicated on each bar (difference in mean is significant with $p <$ 0.05).}
\label{fig:user_success_rate}
\end{figure}

\subsection{Human Evaluation}
We further evaluated the agents that were trained on simulated users against real users, who were recruited from the authors' organization. We conducted a study using the one RL agent and two HRL agents \{RL, h-HRL, m-HRL\}, and compared two pairs: \{RL, m-HRL\} and \{h-HRL, m-HRL\}. In each dialogue session, one agent was randomly selected from the pool to interact with a user. The user was \emph{not} aware of which agent was selected to avoid systematic bias. The user was presented with a goal sampled from a user-goal corpus, then was instructed to converse with the agent to complete the given task.
At the end of each dialogue session, the user was asked to give a rating on a scale from $1$ to $5$ based on the naturalness and coherence of the dialogue; here, $1$ is the worst rating and $5$ the best. In total, we collected $196$ dialogue sessions from $10$ human users.

\begin{figure}[t!]
\centering
\includegraphics[width=\linewidth]{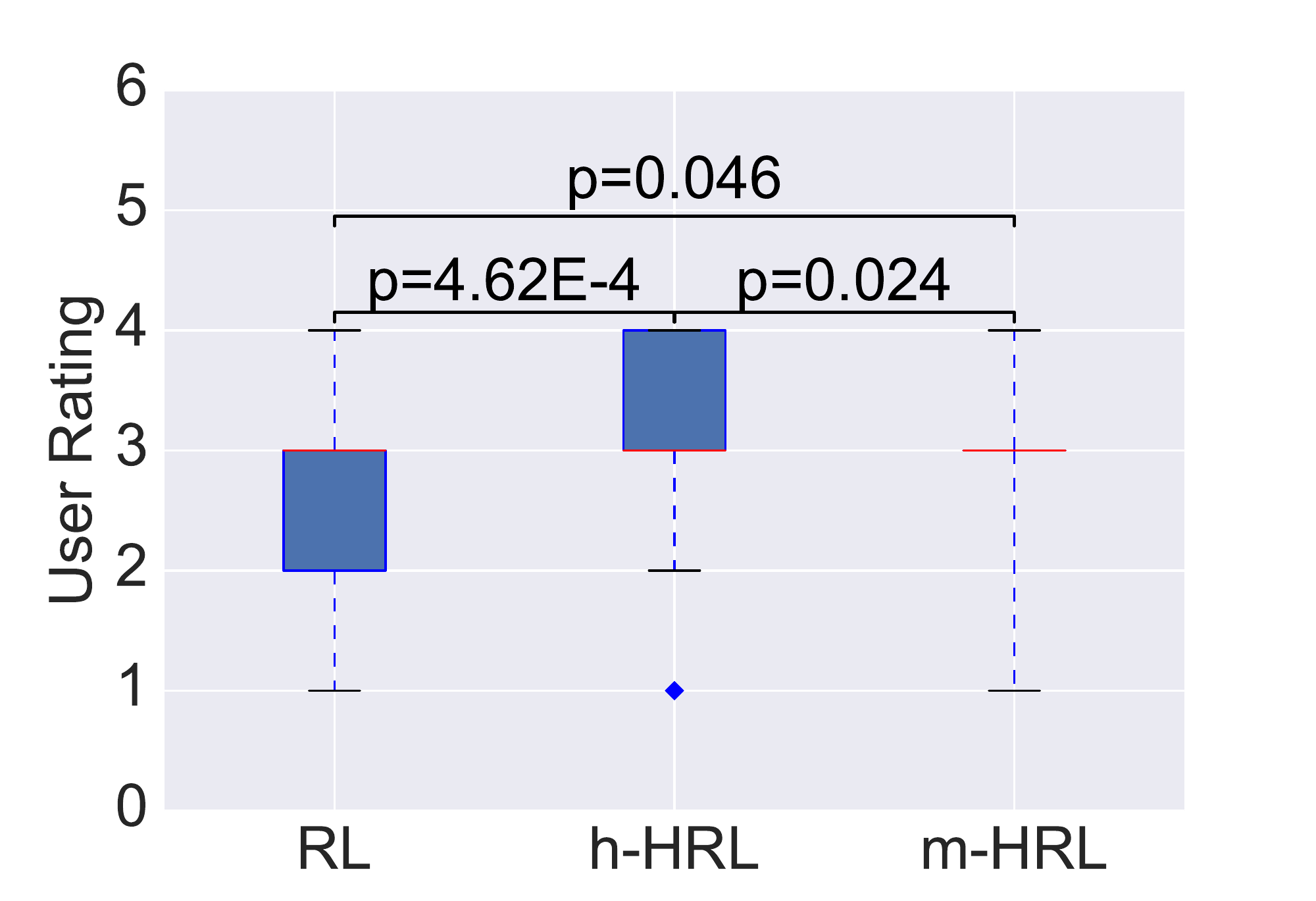}
\vspace{-5mm}
\caption{Distribution of user ratings for three agents in human evaluation}
\label{fig:user_rating}
\end{figure}

Figure~\ref{fig:user_success_rate} summarizes the performances of these agents against real users in terms of success rate. Figure~\ref{fig:user_rating} shows the distribution of user ratings for each agent. For these two metrics, both HRL agents were significantly better than the flat RL agent. Another interesting observation is that the m-HRL agent performs similarly to the h-HRL agent in terms of success rate in the real user study as shown on Figure~\ref{fig:user_success_rate}. Meanwhile in Figure~\ref{fig:user_rating}, the h-HRL agent is significantly better than m-HRL agent in terms of real user ratings. This may be caused by the probabilistic termination of subgoals: we used a threshold strategy to decide whether to terminate a subgoal. This could introduce variance so the agent might not behave reasonably compared with human-defined subgoals which terminate deterministically. 

\subsection{Subgoal Visualization}
\label{sec:subgoal_vis}
Table~\ref{tab:dialog_vis} shows the subgoals discovered by SDN in a sample dialogue by a rule-based agent interacting with the simulated user. The rule-based agent is equipped with a human-defined subtask structure, which always solves subtask \emph{flight} (turns 1--15) before \emph{hotel} (turns 16--23), as shown in the first column. At turn 10, the user starts to talk about \emph{hotel} while the rule-based agent is still working on the pre-defined, unfinished \emph{flight} subtask until subtask \emph{flight} is finished at turn 15. At turn 16, the user switches to \emph{hotel}, and so does the rule-based agent 
until the end of the dialogue. For this rule-based agent, the human-defined subgoal (\emph{flight}) terminates at turn 15. Meanwhile, our SDN model detected two subgoals (except for the final goal): one terminating at turn 9 (Subgoal 1), and another terminating at turn 15 (Subgoal 2).  Subgoal 2 is consistent with the human-defined subgoal. Subgoal 1 is also reasonable since the user tries to switch to \emph{hotel} at turn 10. In Appendix~\ref{sec:app_samples}, Table~\ref{tab:sample_dialogue} shows a sample dialogue session by m-HRL agent interacting with a real user.

\begin{table}[t!]
\scriptsize
\centering
\begin{tabular}[t]{l|l|l}
\hline
h-Task & ID & \multicolumn{1}{c}{Text} \\ \hline
\multirow{16}{*}{\begin{sideways}Flight\end{sideways}} & 1 & \textit{agt}: Hi, may I know your departure city? \\
& 2 & \textit{usr}: I depart from Fort Lauderdale. \\
& 3 & \textit{agt}: And what is your destination city? \\
& 4 & \textit{usr}: I need economy seat to Pittsburgh. \\
& 5 & \textit{agt}: What is your departure date? \\
& 6 & \textit{usr}: 2016-09-12, what time is available for the flight?  \\
& 7 & \textit{agt}: There is one flight at 21:00:00.  \\
& 8 & \textit{usr}: What is the price for that flight?  \\
& 9 & \textit{agt}: \$423. \\
& \multicolumn{2}{c}{\textbox{blue} {Subgoal 1: A detected subgoal terminates!} }  \\
& 10 & \textit{usr}: I need to reserve a hotel for five people.  \\
& 11 & \textit{agt}: What is your return date for the trip?  \\
& 12 & \textit{usr}: 2016-09-18, I need to book 5 tickets for the flight.  \\
& 13 & \textit{agt}: What time do you prefer for return flight?  \\
& 14 & \textit{usr}: What time is available that day?  \\
& 15 & \textit{agt}: There is one flight departing at 11:00:00AM.  \\
\cline{1-1} & \multicolumn{2}{c}{\textbox{blue} {Subgoal 2: A detected subgoal terminates!} }  \\
\multirow{7}{*}{\begin{sideways}Hotel\end{sideways}} & 16 & \textit{usr}: What hotel is available?  \\
& 17 & \textit{agt}: There is one hotel called Winter Hill Inn available.  \\
& 18 & \textit{usr}: What is the price for hotel?  \\
& 19 & \textit{agt}: \$410.  \\
& 20 & \textit{usr}: Great, could you book the trip for me?  \\
& 21 & \textit{agt}: Okay. Trip tickets are issued.  \\
& 22 & \textit{usr}: Thanks very much!  \\
& 23 & \textit{agt}: You are welcome! \\
\hline
\end{tabular}
\caption{Discovered subgoals (except for the final goal) in a sample dialogue by a rule-based agent interacting with user simulator. The left column (h-Task) shows the human-defined subtasks for the rule-based agent. SDN detects two subgoals that terminate at turn 9 and 15 respectively. (h-Task: human-defined subtask, ID: turn ID, \textit{agt}: Agent, \textit{usr}: User)}
\label{tab:dialog_vis}
\end{table}

\section{Related Work}
\label{sec:related_work}
Task-completion dialogue systems have attracted numerous research efforts, and there is growing interest in leveraging reinforcement learning for policy learning. One line of research is on single-domain task-completion dialogues with flat deep reinforcement learning algorithms such as DQN~\cite{zhao2016towards,li2017end,peng2018integrating}, actor-critic~\cite{peng2017adversarial,liu2017iterative} and policy gradients~\cite{williams2017hybrid,liu2017end}. Another line of research addresses multi-domain dialogues where each domain is handled by a separate agent~\cite{gavsic2015policy,gavsic2015distributed,cuayahuitl2016deep}.  Recently, \citet{peng2017composite} presented a composite task-completion dialogue system.  Unlike multi-domain dialogue systems, composite tasks introduce inter-subtask constraints. As a result, the completion of a set of individual subtasks does \textit{not} guarantee the solution of the entire task.

\citet{cuayahuitl10evaluation} applied HRL to dialogue policy learning, although they focus on problems with a small state space.  Later, \citet{budzianowski2017sub} used HRL 
in multi-domain dialogue systems. \citet{peng2017composite} first presented an HRL agent with a global state tracker to learn the dialogue policy in the composite task-completion dialogue systems. All these works are built based on subgoals that were pre-defined with human domain knowledge for the specific tasks. The only job of the policy learner is to learn a hierarchical dialogue policy, which leaves the subgoal discovery problem unsolved. In addition to the applications in dialogue systems, subgoal is also widely studied in the linguistics research community~\cite{allwood2000activity,linell2009rethinking}.  

In the literature, researchers have proposed algorithms to automatically discovery subgoals for hierarchical RL.  One large body of work is based on analyzing the spatial structure of the state transition graphs, by identifying bottleneck states or clusters, among others~\cite{stolle2002learning,mcgovern2001automatic,mannor2004dynamic,simsek05identifying,entezari2011subgoal,bacon2013bottleneck}.  Another family of algorithms identifies commonalities of policies and extracts these partial policies as useful skills~\citep{thrun94finding,pickett02policyblocks,brunskill14pac}.  While similar in spirit to ours, these methods do not easily scale to continuous problems as in dialogue systems.  More recently, researchers have proposed deep learning models to discover subgoals in continuous-state MDPs~\citep{bacon17option,machado17laplacian,vezhnevets17feudal}.  It would be interesting to see how effective they are for dialogue management.


Segmental structures are common in human languages. In the NLP community, some related research on segmentation includes word segmentation~\cite{gao2005chinese,zhang2016transition} to divide the words into meaningful units. Alternatively, topic detection and tracking~\cite{allan1998topic,sun2007topic} segment a stream of data and identify stories or events in news or social text. In this work, we formulate subgoal discovery as a trajectory segmentation problem. Section~\ref{sec:subgoal_discovery} presents our approach to subgoal discovery which is inspired by a probabilistic sequence segmentation model~\cite{wang2017sequence}.

\section{Discussion and Conclusion}
\label{sec:conclusion}
We have proposed the Subgoal Discovery Network to learn subgoals automatically in an unsupervised fashion without human domain knowledge. Based on the discovered subgoals, we learn the dialogue policy for complex task-completion dialogue agents using HRL. Our experiments with both simulated and real users on a composite task of travel planning, show that an agent trained with automatically discovered subgoals performs competitively against an agent with human-defined subgoals, and significantly outperforms an agent without subgoals.
Through visualization, we find that SDN discovers reasonable, comprehensible subgoals given only a small amount of suboptimal but successful dialogue state trajectories.


These promising results suggest several directions for future research. First, we want to integrate subgoal discovery into dialogue policy learning rather than treat them as two separate processes. Second, we would like to extend SDN to identify multi-level hierarchical structures among subgoals so that we can handle more complex tasks than those studied in this paper. Third, we would like to generalize SDN to a wide range of complex goal-oriented tasks beyond dialogue, such as the particularly challenging Atari game of Montezuma's Revenge~\cite{kulkarni2016hierarchical}. 

\section*{Acknowledgments}
We would like to thank the anonymous reviewers, members of the xlab at the University of Washington, and Chris Brockett, Michel Galley for their insightful comments on the work. Most of this work was done while DT, CW \& LL were with Microsoft.

\bibliography{emnlp2018}
\bibliographystyle{acl_natbib_nourl}


\appendix



\section{Hierarchical Dialogue Policy Learning}
\label{app:hrl}
This section provides more algorithmic details for Section~\ref{sec:policy_learning}.  Again, assume a conversation of length $T$:
\[
\tau = (s_0,a_0,r_0,\ldots,s_{T-1},a_{T-1},r_{T-1},s_T)\,.
\]
Suppose an HRL agent segments the trajectory into a sequence of subgoals as $g_0, g_1,\ldots\in\mathcal G$, and the corresponding subgoal termination time steps as $t_{g_0},t_{g_1},\ldots\in\mathbb N^*$.
Furthermore, denote the intrinsic reward at time step $t$ by $r^i_t$.
%
%
The top-level and low-level Q-functions satisfy the following Bellman equations:
\begin{equation}
\begin{aligned}
Q^*(s,g)=\mathbb E\Big[ &\sum\limits_{i=t_{g_j}}^{t_{g_{j+1}}-1}\gamma^{i-t_{g_j}} r^e_{t+1} \\
&+\gamma^{t_{g_{j+1}}-t_{g_j}} \cdot\max\limits_{g'\in\mathcal G}Q^*(s_{t_{g_{j+1}}},g')\\
			&|s_{t_{g_j}}=s, g_j=g\Big] \nonumber
\end{aligned}
\label{eqn:re}
\end{equation}
and
\begin{equation}
\begin{aligned}
Q^*(s,a,g)=\mathbb E\Big[ & r^i_{t}+\gamma\cdot\max\limits_{a'\in\mathcal A}Q^*_i(s_{t+1},g_j,a')\\
 &|s_t=s, g_j=g, a_t=a, \\
 & t\in[t_{g_j}, t_{g_{j+1}})\Big]\,.\nonumber
\end{aligned}
\label{eqn:ri}
\end{equation}
Here $\gamma\in[0, 1]$ is a discount factor, and 
the expectations are taken over the randomness of the reward and the state transition, 

We use deep neural networks to approximate the two Q-value functions as $Q^*(s,a,g)\approx Q(s,a,g;\theta_i)$ and $Q^*(s,g)\approx Q(s,g;\theta_e)$.  The parameters $\theta_i$ and $\theta_e$ are optimized to minimize the following quadratic loss functions:
\begin{equation}
\begin{aligned}
L_i(\theta_i)=&\frac{1}{2|D^i|}\sum_{(s,a,g,s',r^i)\in\mathcal D^i}[(y^i-Q(s,a,g;\theta_i))^2]\\
y^i=&r^i+\gamma\cdot\max\limits_{a'\in\mathcal A}Q_i(s',a',g;\theta_i)
\end{aligned}
\label{eqn:loss-i}
\end{equation}
and 
\begin{equation}
\begin{aligned}
L_e(\theta_e)=&\frac{1}{2|D^e|}\sum_{(s,g,s',r^e)\in\mathcal D^e}[(y^e-Q(s,g;\theta_e))^2]\\
y^e=&r^e+\gamma\cdot\max\limits_{g'\in\mathcal G}Q(s',g';\theta_e)\,.
\end{aligned}
\label{eqn:loss-e}
\end{equation}
Here, $\mathcal D^e$, $\mathcal D^i$ are the replay buffers storing dialogue experience for training top-level and low-level policies.

Optimization of parameters $\theta_i$ and $\theta_e$ can be done by stochastic gradient descent on the two loss functions in Equations~\eqref{eqn:loss-i} and \eqref{eqn:loss-e}. The gradients of the two loss functions w.r.t their parameters are
\begin{equation}
\begin{aligned}
\nabla_{\theta_i} L_i=\frac{1}{|D^i|}\sum_{(s,a,g,s',r^i)\in\mathcal D^i}&\Big[\nabla_{\theta_i}Q(s,a,g;\theta_i)\cdot\\
&(y^i-Q_i(s,a,g;\theta_i))\Big] \nonumber
\end{aligned}
\end{equation}
and
\begin{equation}
\begin{aligned}
\nabla_{\theta_e}L_e=\frac{1}{|D^e|} \sum_{(s,g,s',r^e)\in\mathcal D^e}&\Big[\nabla_{\theta_e}Q(s,g;\theta_e)\cdot\\
&(y^e-Q_e(s,g;\theta_e))\Big]\,. \nonumber
\end{aligned}
\end{equation}
To avoid overfitting, we also add $L_2$-regularization to the objective functions above.

\newpage

\section{Sample Dialogue}
\label{sec:app_samples}

\begin{table}[ht!]
\footnotesize
\centering
\vspace{-2mm}
\caption{Sample dialogue by the m-HRL agent interacting with real user: bolded slots are the joint constraints between two subtasks. (\textit{agt}: Agent, \textit{usr}: User)}
\begin{tabular}[t]{ll}
\hline
\multicolumn{2}{c}{\textbf{User Goal}} \\
\begin{tabular}[ct]{@{}l@{}}
\textit{reserve-hotel} subtask:\\
\{\\
\-\hspace{0mm} ``request\_slots": \{	 \-\hspace{10mm}     ``inform\_slots": \{\\
\-\hspace{0mm}    ``hotel\_price": ``?"	 \-\hspace{10mm}    \textbf{``hotel\_date\_checkin":``2016-09-22"} \\
\-\hspace{0mm}    ``hotel\_date\_checkout": ``?"	\-\hspace{-1mm}    \textbf{``hotel\_city": ``Curitiba"}	\\
\-\hspace{0mm}    ``hotel\_name": ``?"	 \-\hspace{10mm}      \textbf{``hotel\_numberofpeople": ``4"} \\
\-\hspace{0mm}    ``hotel\_amenity\_wifi": ``?"	 \-\hspace{1mm}       \}\\
\-\hspace{0mm}  \}	 \\
\} \\
\end{tabular}
\\
\\
\begin{tabular}[ct]{@{}l@{}}
\textit{book-flight-ticket} subtask:\\
\{\\
\-\hspace{0mm} ``request\_slots": \{ \-\hspace{8mm}     ``inform\_slots": \{\\
\-\hspace{0mm}    ``price": ``?"	\-\hspace{15.5mm}    ``or\_city": ``Lima",	\\
\-\hspace{0mm}    ``return\_time\_dep": ``?"	\-\hspace{2mm}   \textbf{``dst\_city": ``Curitiba"},\\
\-\hspace{0mm}    ``return\_date\_dep": ``?"	\-\hspace{2mm}   \textbf{``numberofpeople": ``4"},\\
\-\hspace{0mm}    ``depart\_time\_dep": ``?"	\-\hspace{2mm}    \textbf{``depart\_date\_dep":``2016-09-22"}	\\
\-\hspace{0mm}    ``seat": ``?"	\-\hspace{17mm}  \}\\
\-\hspace{0mm}  \}	\\
\}\\
\end{tabular} \\
\hline
\multicolumn{2}{c}{\textbf{m-HRL agent}} \\
\multicolumn{2}{l}{\textit{agt}: Hi, may I know your departure city and destination city?} \\
\multicolumn{2}{l}{\textit{usr}: I depart from Lima to Curitiba.} \\
\multicolumn{2}{l}{\textit{agt}: What is your departure date?} \\
\multicolumn{2}{l}{\textit{usr}: I am leaving on 2016-09-22.}  \\
\multicolumn{2}{l}{\textit{agt}: There is one flight available at 8:00:00AM.}  \\
\multicolumn{2}{l}{\textit{usr}: What date is available for return flight?}  \\
\multicolumn{2}{l}{\textit{agt}: There is one return flight on 2016-09-27.}  \\
\multicolumn{2}{l}{\textit{usr}: What is the time for the return flight?}  \\
\multicolumn{2}{l}{\textit{agt}: It is 00:00:00AM.} \\
\multicolumn{2}{l}{\textit{usr}: What seat is available?}  \\
\multicolumn{2}{l}{\textit{agt}: Only Economy seat now.}  \\
\multicolumn{2}{l}{\textit{usr}: We need 4 tickets for the flight.}  \\
\multicolumn{2}{l}{\textit{agt}: Total is 1201 dollars for flight.}  \\
\multicolumn{2}{c}{\textbox{blue} {inform(subtask\_complete): Subgoal reached!}}  \\
\multicolumn{2}{l}{\textit{usr}: What hotel is available there?}  \\
\multicolumn{2}{l}{\textit{agt}: There is one hotel called Starlight.}  \\
\multicolumn{2}{l}{\textit{usr}: Is the wifi available for the hotel?}  \\
\multicolumn{2}{l}{\textit{agt}: Yes, it is free.}  \\
\multicolumn{2}{l}{\textit{usr}: What's the price for the hotel?}  \\
\multicolumn{2}{l}{\textit{agt}: It is 1541 dollars for hotel.}  \\
\multicolumn{2}{l}{\textit{usr}: Could you please book the trip for me?}  \\
\multicolumn{2}{l}{\textit{agt}: Okay. Trip tickets are issued.}  \\
\multicolumn{2}{l}{\textit{usr}: Thanks!}  \\
\multicolumn{2}{l}{\textit{agt}: Thanks!} \\
\hline
\end{tabular}
\label{tab:sample_dialogue}
\vspace{-2mm}
\end{table}

\end{document}

%% file: section_hrl.tex
\subsection{Hierarchical Dialogue Policy Learning}
\label{sec:policy_learning}
Before describing how we use a trained SDN model for HRL, we first present a short review of HRL for a task-oriented dialogue system.
%
%
Following the \emph{options} framework~\cite{sutton1999between}, assume that we have a state set $\mathcal S$, an option set $\mathcal G$ and a finite primitive action set $\mathcal A$.

The HRL approach we take learns two Q-functions~\cite{peng2017composite}, parameterized by $\theta_e$ and $\theta_i$, respectively:
\begin{itemize}[noitemsep,leftmargin=*,topsep=0pt]
\item{The top-level $Q^*(s,g;\theta_e)$ measures the maximum total discounted \emph{extrinsic} reward received by choosing subgoal $g$ in state $s$ and then following an optimal policy. These extrinsic rewards are the objective to be maximized by the entire dialogue policy.}
\item{The low-level $Q^*(s,a,g;\theta_i)$ measures the maximum total discounted intrinsic reward received to achieve a \emph{given} subgoal $g$, by choosing action $a$ in state $s$ and then following an optimal option policy. These intrinsic rewards are used to learn an option policy to achieve a given subgoal.}
\end{itemize}

\vspace{2mm}
Suppose we have a dialogue session of $T$ turns: $\tau=(s_0,a_0,r_0,\ldots,s_T)$, which is segmented into a sequence of subgoals $g_0, g_1, \ldots \in \mathcal G$.  Consider one of these subgoals $g$ which starts and ends in steps $t_0$ and $t_1$, respectively.

The top-level Q-function is learned using Q-learning, by treating subgoals as temporally extended actions:
\[
\theta_e \leftarrow \theta_e + \alpha \cdot \left(q - Q(s_t,g;\theta_e)\right) \cdot \nabla_{\theta_e} Q(s_t,g;\theta_e)\,,
\]
where
\[
q = \sum_{t=t_0}^{t_1-1} \gamma^{t-t_0} r_t^e + \gamma^{t_1-t_0} \max_{g'\in\mathcal G} Q(s_{t_1},g';\theta_e)\,,
\]
and $\alpha$ is the step-size parameter, $\gamma\in[0, 1]$ is a discount factor. In the above expression of $q$, the first term refers to the total discounted reward during fulfillment of subgoal $g$, and the second to the maximum total discounted after $g$ is fulfilled.

The low-level Q-function is learned in a similar way, and follows the standard Q-learning update, except that intrinsic rewards for subgoal $g$ are used. Specifically, for $t = t_0, t_0+1,\ldots, t_1-1$: \vspace{-2mm}

{\small
\[
\theta_i \leftarrow \theta_i + \alpha \cdot \left(q_t - Q(s_t,a_t,g;\theta_e)\right) \cdot \nabla_{\theta_i} Q(s_t,a_t,g;\theta_i)\,,
\]}
where
\[
q_t = r_t^i + \gamma \max_{a' \in \mathcal A} Q(s_{t+1},a',g;\theta_i)\,.
\]
Here, the intrinsic reward $r_t^i$ is provided by the internal critic of dialogue manager. More details are in Appendix~\ref{app:hrl}.

In hierarchical policy learning, the combination of the extrinsic and intrinsic rewards is expected to help the agent to successfully accomplish a composite task as fast as possible while trying to avoid unnecessary subtask switches.  Hence, we define the extrinsic and intrinsic rewards as follows: 

\paragraph{Extrinsic Reward.} Let $L$ be the maximum number of turns of a dialogue, and $K$ the number of subgoals. At the end of a dialogue, the agent receives a positive extrinsic reward of $2L$ for a success dialogue, or $-L$ for a failure dialogue; for each turn, the agent receives an extrinsic reward of $-1$ to encourage shorter dialogues.

\paragraph{Intrinsic Reward.} When a subgoal terminates, the agent receives a positive intrinsic reward of $2L/K$ if a subgoal is completed successfully, or a negative intrinsic reward of $-1$ 
otherwise; for each turn, the agent receives an intrinsic reward $-1$ to encourage shorter dialogues.
